\documentclass[sigconf]{acmart}

\usepackage{booktabs} % For formal tables

\usepackage{subfigure}
\usepackage{graphicx}
\usepackage{multirow}
\usepackage{color}
\usepackage{soul}
\usepackage{geometry}
\usepackage{algorithm}
\usepackage{algorithmic}
\usepackage{kotex}

\usepackage{pifont}

\usepackage{hyperxmp}   % 반드시 hyperref 바로 뒤에
\makeatother

\copyrightyear{2025} 
\acmYear{2025} 
\setcopyright{acmlicensed}\acmConference[arXiv]{}{Aug 27, 2025}{NCSOFT}
\acmBooktitle{arXiv}
\begin{document}
\title{Human-AI Collaborative Bot Detection in MMORPGs}

\author{Jaeman Son}
\authornote{Equal contribution}
\orcid{1234-5678-9012}
\affiliation{%
  \institution{NCSOFT}  
  \country{Republic of Korea}
}
\email{jaemanson@ncsoft.com}

\author{Hyunsoo Kim}
\authornotemark[1]
\orcid{1234-5678-9012}
\affiliation{%
  \institution{NCSOFT}  
  \country{Republic of Korea}
}
\email{aitch25@ncsoft.com}

\newcommand\tc{black}
\newcommand{\hs}[2][]{\st{#1}\color{black}#2 \color{black}}
\newcommand{\ck}[2][]{\st{#1}\color{black}#2 \color{black}}
\newcommand{\rv}[2][]{\st{#1}\color{black}#2 \color{black}}

\begin{abstract}

In Massively Multiplayer Online Role-Playing Games (MMORPGs), auto-leveling bots exploit automated programs to level up characters at scale, undermining gameplay balance and fairness. Detecting such bots is challenging, not only because they mimic human behavior, but also because punitive actions require explainable justification to avoid legal and user experience issues.
In this paper, we present a novel framework for detecting auto-leveling bots by leveraging contrastive representation learning and clustering techniques in a fully unsupervised manner to identify groups of characters with similar level-up patterns. 
To ensure reliable decisions, we incorporate a Large Language Model (LLM) as an auxiliary reviewer to validate the clustered groups, effectively mimicking a secondary human judgment. We also introduce a growth curve-based visualization to assist both the LLM and human moderators in assessing leveling behavior.
This collaborative approach improves the efficiency of bot detection workflows while maintaining explainability, thereby supporting scalable and accountable bot regulation in MMORPGs.

\end{abstract}




\maketitle

\section{Introduction}
\label{sec: Introduction}

In MMORPGs (Massively Multiplayer Online Role-Playing Games), bots exhibit intelligent and systematic behavior by automating character progression along optimized routes. They efficiently farm experience and items by targeting high-yield areas and completing quests, eventually forming farming units—typically composed of three or more characters—to acquire valuable resources. These automated actions result in far greater efficiency than human players, disrupting fair play and destabilizing the in-game economy through item monopolization and real money trading (RMT) \cite{huhh2008simple,lee2018no,kwon2016crime,lee2011detecting}.

Effective bot detection is essential, but in practice, sanctioning often leads to legal disputes—especially when evidence is insufficient or legitimate players are wrongly penalized \cite{kim2025framework}. Thus, detection systems must ensure both accuracy and explainability.

This paper focuses on detecting auto-leveling bots, a specific class of game bots, by leveraging time-series representation models and Large Language Models (LLMs). Our framework builds upon prior approaches such as \cite{kim2025framework,tao2018nguard,xu2020nguard+}. 
We construct sequential inputs from each character’s level-up logs and temporal features, and project them into a latent space through a time-series representation model. The embedded vectors are then clustered using DBSCAN \cite{ester1996density}, a density-based clustering algorithm well-suited for discovering arbitrarily shaped groups. This fully unsupervised process incurs zero labeling cost and allows bots—whose leveling behavior is highly systematic—to form dense clusters, while human players with irregular progression patterns remain isolated.

To ensure reliability, we visualize clustered groups using growth curve-based plots and leverage LLMs for secondary verification to assess machine-like behavior. While our model captures systematic patterns well, ambiguous cases may still arise when bot-like and human-like behaviors are similar. To address this, we delegate verification to LLMs, reducing manual effort and enabling human moderators to focus on higher-level decisions.

The contributions of this paper are as follows:
\begin{itemize}
\item We propose the first detection framework for auto-leveling bots in MMORPGs that operates in a fully unsupervised manner—requiring no labeled data—and supports intuitive \hs{and interpretable} analysis through \hs{level-up interval visualizations} of detected bot groups.
\item To enhance the reliability of detection results, we further incorporate Large Language Model (LLM)-based secondary verification. This novel component sheds light on the decision-making process behind unsupervised detection and facilitates accountable bot regulation.
\end{itemize}

\section{Related works}
\label{sec: Related works}

\subsection{Bot detection tasks}

The task of game bot detection varies by genre \cite{chen2008game,pao2010game,pinto2021deep,kanervisto2022gan,choi2023botscreen,lee2018no,qi2022gnn,su2022trajectory,tao2019mvan,zhao2022t,kim2025framework,lee2016you,kang2016multimodal,pu2022unsupervised}. Particularly, MMORPGs face a threat of large-scale bot farms driven by automation. These bots prioritize efficiency and profit over competition, using optimized scripts. Detecting such bots has been the focus of several studies.

Prior work has explored diverse features for detecting abnormal behavior in MMORPGs, such as status logs, event records, quest and trade histories, click \hs{(touch)} patterns, movement paths, and self-similarity metrics \cite{lee2018no,qi2022gnn,su2022trajectory,tao2019mvan,zhao2022t,kim2025framework,lee2016you,kang2016multimodal,pu2022unsupervised}.

% These studies can be grouped into two categories. The first aims to minimize human intervention in operations \cite{qi2022gnn,su2022trajectory,tao2019mvan,zhao2022t,lee2016you,kang2016multimodal,pu2022unsupervised}, using multimodal inputs to enhance model performance. However, these approaches often result in complex models that hinder effective human–model interaction—an issue in practice, since accurate bot regulation depends on human oversight to avoid false positives and legal risks.
\rv{
Many recent approaches aim to reduce human intervention by leveraging multimodal inputs to enhance model performance \cite{qi2022gnn,su2022trajectory,tao2019mvan,zhao2022t,lee2016you,kang2016multimodal,pu2022unsupervised}. While these methods can improve automation, they often result in complex models that hinder effective human–model interaction. This presents a practical challenge, as accurate bot regulation still requires human oversight to prevent false positives and mitigate legal risks.
}

To address this, \cite{kim2025framework} introduced BotTRep, a framework that facilitates interaction between human experts and models through intuitive and explainable materials to reduce false positives. Building on this philosophy, our study positions the model as a supportive tool for both LLMs and game masters. We demonstrate the feasibility of replacing the labor-intensive verification process—previously handled by humans in BotTRep—with LLM-based automation.

\subsection{LLM-assisted verification}

Recently, there has been a growing interest in leveraging Large Language Models (LLMs) for time series tasks \cite{gruver2023large,jin2023time, dong2024can,liu2024large,chow2024towards,merrill2024language,xie2024chatts,zhang2025tempogpt,kong2025time}. In particular, we focus on recent efforts that utilize the general intelligence of LLMs \hs{in zero- or few-shot settings,} without requiring task-specific training \cite{gruver2023large,jin2023time,dong2024can,liu2024large}.
Beyond time series applications, LLMs have also been actively explored in a wide range of judgment-based tasks, where they have in some cases demonstrated accuracy comparable to human-level decision making \cite{liu2024interpretable,gu2024survey,zheng2023judging}.

Particularly, as natural language models, LLMs not only make judgments on time series tasks, but also provide explanations for their decisions in natural language, facilitating effective human interaction. In this paper, we leverage the strengths of LLMs to enhance the reliability of game bot detection results. Specifically, we design a data flow where suspicious clusters detected by the auto-leveling bot model undergo secondary verification by the LLM. This ensures that the final output delivered to the game master has been double-checked, thereby reducing the risk of false positives.

\section{Proposed Approach}
\label{sec: Proposed Approach}

\begin{figure*}[t]
\centering
\includegraphics[width=\textwidth]{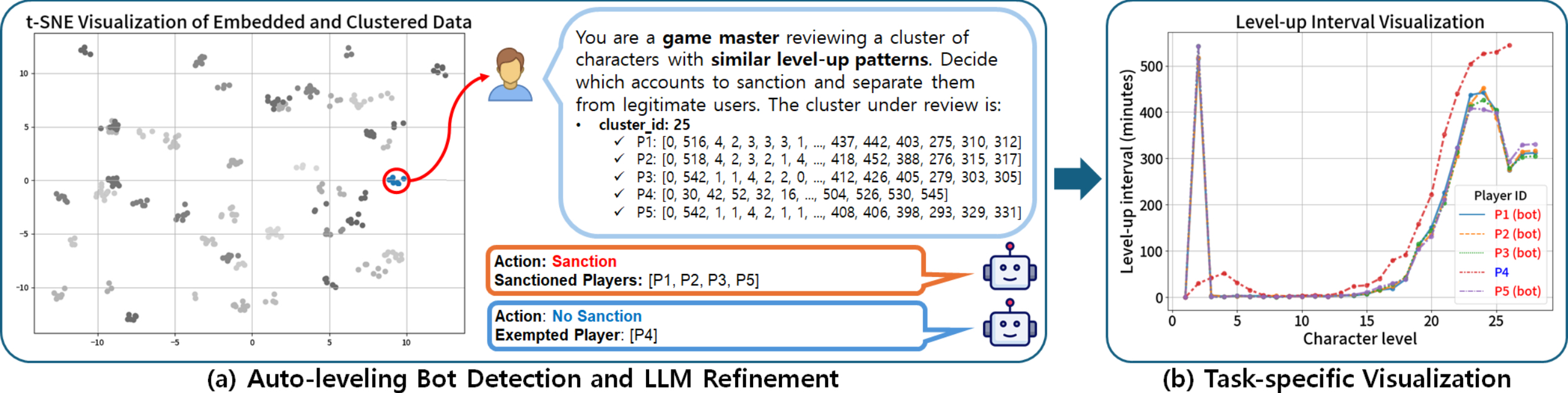} 
\vskip -4mm
\caption{This figure summarizes the core mechanism of our framework. It begins by embedding level-up interval sequences using a time-series representation model. Bots with similar sequences are expected to yield similar embeddings, which are then clustered via DBSCAN. As the framework is fully unsupervised to reduce labeling costs, we verify—using the center chart—whether normal users are mistakenly included in clusters. To automate this previously manual verification, we incorporate LLMs into the process.}
\label{fig:overview}   
\vskip -4mm
\end{figure*}

This section presents our framework components, with the core mechanism shown in Figure~\ref{fig:overview}.

\subsection{Data description}

\subsubsection{Data preperation}

We use time-series data collected from an MMORPG, where each record indicates a character's level-up event with a timestamp. For each character \( p \in \mathbf{P} \), we define the level-up time sequence as \( T^{(p)} = \{t_1, \ldots, t_i, \ldots, t_{\min(50, l^{(p)})}\} \), where \( l^{(p)} \) denotes the highest level reached, and \( t_i \) indicates the time (in minutes) it took to reach level \( i \) from level \( i-1 \).

To ensure consistency, we cap all sequences at level 50, since higher levels often involve irregular progression patterns due to PvP or social interactions, even for bots. This threshold provides a stable basis for automated behavior analysis.

\hs{In this study, we only conduct experiments on cases where character level-ups were properly logged. Characters with missing data or level-up logs influenced by paid items were excluded, as their progression could not be reliably observed.}

% \subsubsection{Excluded data}

% When constructing the dataset, we excluded cases in which experience potions were used to level up rapidly, as well as cases where level-up logs were not properly recorded. The rationale for this exclusion is that our model is designed to detect automated character training based on behavioral data accumulated during the character development process.

% In cases where characters level up using experience potions, the training behavior at each level is often insufficiently logged, which may impair the model’s decision-making accuracy. In other words, our model is designed to prioritize high precision.

% This reflects a strategic choice to minimize the risk of false positives—i.e., misidentifying normal users as automated ones—which can lead to negative user experiences or even legal disputes during service operations. Our detection framework focuses on minimizing false positives rather than false negatives, and accordingly, we excluded cases with missing or incomplete logs from the analysis.

\subsection{Auto-leveling bot detection model}

\subsubsection{Represntation model}
% The model used in this study takes as input the sequence data \( T^{(p)} \), which consists of the numerical values defined earlier, and extracts a representation from it. Specifically, the model is designed to produce similar representations for similar time-series sequences, and dissimilar representations for different sequences. Any model that adheres to this format is applicable within our framework.

The model used in this study takes as input the sequence data \( T^{(p)} \), which consists of the numerical values defined earlier, and extracts a representation from it. Since the sequence length varies by character, the model must be capable of handling variable-length inputs and producing appropriate representations accordingly. Specifically, the model is designed to generate similar representations for similar time-series sequences, and dissimilar ones for different sequences. Any model that meets these requirements can be applied within our framework.

In this study, we primarily used the TS2Vec model \cite{yue2022ts2vec}. However, we modified the original implementation to output only the representation corresponding to the input sequence. The process of extracting the representation vector from the model can be simply expressed as \( r^{(p)} = \mathcal{M}(T^{(p)}) \).

The rationale behind using a representation model lies in the observation that auto-leveling bots tend to follow highly optimized leveling routes, frequently repeating behaviors such as visiting the same hunting grounds or completing the same quests. These patterns result in highly consistent numerical signals, which are captured as closely located vectors in the latent space. In contrast, human players typically exhibit greater variation due to unstructured gameplay, resulting in more dispersed representations.

\subsubsection{Clustering algorithm}

The clustering algorithm described in this section performs clustering on these representation vectors, and we primarily adopt DBSCAN \cite{ester1996density} for this task. To use DBSCAN effectively, two key parameters must be configured: \( min\_sample \) and \( \varepsilon \). DBSCAN clusters data points when at least \( min\_sample \) points are within a distance of \( \varepsilon \) from one another; otherwise, the points are treated as noise. The clustering process in our framework is formalized as \( c^{(p)} = \mathbf{cluster}(r^{(p)}) \), where \( r^{(p)} \) is the representation of character \( p \).

In our study, we set \( min\_sample = 3 \), based on the observation that bot activity in the field typically occurs in groups of three or more. The parameter \( \varepsilon \) was determined following the method proposed in BotTRep \cite{kim2025framework}, which is based on the adaptive density estimation described in \cite{schubert2017dbscan}.
Specifically, we \hs{adopted} \( \varepsilon = \text{quantile}_{q}(dist) \), where \( dist \) refers to the distance between each data point and its \( k \)th nearest neighbor. In our experiments, we set \( q \in \{0.1, 0.2\} \).

\subsection{LLM-assisted verification}

\subsubsection{LLM-based Verification Module}

After applying the clustering step, we incorporated a Large Language Model (LLM) to refine the results by filtering out potential false positives. Specifically, we provided the LLM with a list of character groups that had been pre-clustered by our model and tasked it with verifying whether all characters within each cluster were indeed bots. If any non-bot character was detected within a cluster, the character was excluded from the final list of bot candidates.

This task originally required manual inspection of leveling curves at the cluster level by human operators. However, this process is highly repetitive and labor-intensive. To address this, we designed our system to offload the task to an LLM. The LLM receives a predefined prompt along with the original time interval sequences \( T^{(p)} \) of the characters in each cluster as input, and performs verification accordingly, \hs{as shown in Figure~\ref{fig:overview}.}

This process is defined as \( \mathcal{B} = \mathbf{LLM}(\mathcal{T}, \mathcal{C}, pt) \), where \( T^{(p)} \in \mathcal{T} \) and \( c^{(p)} \in \mathcal{C} \). Here, \( T^{(p)} \) denotes the original time interval sequences, $c^{(p)}$ represents the clustering result, and \( pt \) is the task-specific prompt provided to guide the LLM. In this study, we used GPT-4o for our experiments.

To address this task, we employed a hybrid approach that integrates a time-series model with an LLM, rather than relying exclusively on an LLM-based solution. Although we also evaluated GPT-4o by providing raw level-up logs as input, this configuration suffered from input length limitations and model constraints, resulting in unreliable outputs.

\subsubsection{Prompt engineering}

We constructed prompts for the LLM based on the following strategy: 1) role assignment, 2) definition of criteria for determining whether a character should be sanctioned, and 3) input–output format design. As discussed earlier, the LLM is used to help filter out legitimate users who were incorrectly included in clusters. When normal users—who ideally should not be clustered—are grouped together, the LLM analyzes their level-up interval sequences in a zero-shot manner to determine whether they should be excluded from the list of sanction candidates. This task includes diverse cases where bots and legitimate users are mixed. To avoid potential bias from showing only a few samples in a few-shot setting, we adopted a zero-shot approach \hs{by providing explicit criteria within the prompt instead of giving specific examples.} 

In step 2), we instructed the LLM to distinguish between auto-leveled bots and individually-leveled legitimate characters. To improve performance, we adopted the Chain-of-Thought (CoT) prompting approach \cite{wei2022chain,liu2024logprompt}, guiding the model through the following steps: a) understand the input, b) compare level-up intervals, c) check structural similarity, d) identify bot groups, e) exclude non-bot characters, and f) produce the final output. The LLM receives level-up interval data of characters, organized in cluster-based batches, as text input.

% We modeled the prompt in a zero-shot manner, as few-shot prompting resulted in degraded performance. This is likely due to the wide variety of legitimate user behaviors that were incorrectly clustered; providing limited examples in few-shot format led the model to become biased toward specific patterns.

\subsection{Level-up interval visualization}

We propose a level-up interval visualization method to illustrate the outcomes of clustering and LLM-based refinement. This method sequentially shows the time (in minutes) taken to progress from one level to the next.

% It helps identify false positives by visually highlighting legitimate users mistakenly grouped with bots. When clustering is accurate, characters in a cluster exhibit nearly identical leveling curves, while normal users appear noticeably different.

It helps identify false positives by visually highlighting legitimate users mistakenly grouped with bots, \hs{enhancing the interpretability of the detection results}. When clustering is accurate, characters in a cluster exhibit nearly identical leveling curves, while normal users appear noticeably different.

In practice, the game master can use it to verify whether legitimate users are properly excluded before finalizing sanctions, as shown in Figure~\ref{fig:overview}~(b).

\section{Experiments}
\label{sec: Experiments}

To systematically assess the effectiveness of the proposed approach, we carry out three types of evaluation.

\subsection{Dataset}

% 익명화 처리 -- 리뷰 이후 복원 예정
% In this study, we conduct experiments using data from three mobile MMORPGs published by NCSOFT: Lineage M (LM), Lineage 2M (L2M), and Lineage W (LW). 

% In this study, we conduct experiments using data from three large-scale mobile MMORPGs operated by a commercial game publisher. Each game has an average daily active user base of approximately 200K, reflecting the scale and commercial relevance of the dataset. For model training, we use approximately three months of gameplay data collected from October 1, 2024, to December 31, 2024. For evaluation, we utilize a separate dataset spanning January 1, 2025, to January 14, 2025, covering about two weeks.

% Notably, the evaluation is conducted not on the entire set of game servers, but on the three most recently opened worlds (as of the data collection date) from each of the G1, G2, and G3\footnote{Each game title was anonymized for review and will be disclosed upon publication.} titles. The total number of data points used for training is 1,005,522, and the number used for evaluation is 38,514.

In this study, we use data from three large-scale mobile MMORPGs operated by a commercial game publisher, each with approximately 200K daily active users. For training, we use three months of gameplay logs (October 1 to December 31, 2024), and for evaluation, a separate two-week dataset (January 1 to 14, 2025). \hs{The training set consists of 1,005,522 data points, and the evaluation set includes 38,514 data points.} Evaluation is performed on the three most recently opened worlds (as of the data collection date) from each anonymized title: G1, G2, and G3\footnote{Each game title was anonymized for review and will be disclosed upon publication.}.

\subsection{Evaluation of embedding quality}

\subsubsection{Evaluation metric}
\hs{The first evaluation assesses the embedding quality of the representation model. A better model more clearly separates bots, which should be clustered, from legitimate users, which should not.}

To evaluate this, we generate ten perturbed variants of each character's level-up interval sequence: $T^{(p)}_\text{pert} = T^{(p)} + \mathcal{N}_{lv}$.

Specifically, we implement a more realistic perturbation pipeline composed of three sequential operations: random deletion, additive perturbation applied with probability, and index-wise swapping. We first apply random deletion to the original sequence $T^{(p)}$: $T^{(p)}_{\text{del}, lv} = [t_i \in T^{(p)} \mid u_i > p_{\text{del}},\ u_i \sim \text{Uniform}(0,1)],\ p_{\text{del}} = 0.05 \times lv$.

Then, we construct the perturbed sequence with conditional additive noise:
\[
T^{(p)}_{\text{pert}, lv} = T^{(p)}_{\text{del}, lv} + \mathcal{N}_{lv}, \ \ \mathcal{N}_{lv} \sim 
\begin{cases}
\text{Uniform}(-3 \cdot lv,\; 3 \cdot lv), & \text{w.p. } \frac{lv}{10} \\
0, &\text{otherwise}
\end{cases}
\]

Finally, we apply index-wise swapping to further distort the temporal order: $\text{pert}^{(p)}_{lv} = \text{Swap}\left( T^{(p)}_{\text{pert}, lv},\ \{(a_k, b_k)\}_{k=1}^{lv} \right)$ where $\text{Swap}(\cdot)$ denotes the \textit{index-wise swapping} operation applied to $lv$ randomly selected index pairs, and $a_k$ and $b_k$ are the indices selected for swapping. That is, when $lv = 1$, one swap is performed; when $lv = 2$, two swaps are performed, and so on.

This staged perturbation introduces increasing levels of structural corruption as $lv$ increases, enabling a controlled evaluation of representation robustness. Ideally, as \( lv \) increases, the perturbed sequence \( \text{pert}^{(p)}_{lv} \) diverges further from the original sequence \( T^{(p)} \). Thus, a well-functioning representation model should satisfy the following condition:

\begin{align}
d\big(\mathcal{M}(T^{(p)}),\ \mathcal{M}(\text{pert}^{(p)}_{i})\big)
< d\big(\mathcal{M}(T^{(p)}),\ \mathcal{M}(\text{pert}^{(p)}_{j})\big),& \nonumber \\
\text{for } 1 \leq i < j \leq 10&
\label{eq:pert_alignment}
\end{align}
where \( d(\cdot, \cdot) \) is the Euclidean distance. To assess how well the structure in Equation~(\ref{eq:pert_alignment}) is preserved, we use Kendall's Tau~\cite{kendall1938new}, which gives higher scores when the alignment is better maintained.

\subsubsection{Baseline experiment}

To identify the most suitable representation model for our dataset, we conducted a baseline experiment using three representative approaches. The first model is Dynamic Time Warping (DTW), a fundamental method for time-series comparison. The second is an Autoencoder, commonly used to extract representations from various types of sequential data. The third is TS2Vec, which has been reported to provide universal time-series representations.
Among the three, TS2Vec demonstrated the best performance in our baseline experiment, as summarized in Table~\ref{tab:eval1}.

% \textcolor{red}{In addition, we also explored using GPT-4o by directly feeding it raw level-up logs of all characters. However, this approach encountered significant limitations due to excessive input length and model constraints, resulting in unreliable outputs. Therefore, we excluded this configuration from the baseline comparison table.}

\begin{table}[H]
\begin{tabular}{cccc}
\hline
Model         & DTW    & Autoencoder & TS2Vec \\ \hline
Kendall's Tau & 0.5283 & 0.8219      & \textbf{0.8304} \\ \hline
\end{tabular}
\caption{TS2Vec showed the best embedding quality based on Kendall's Tau.}
\label{tab:eval1}
\vskip -8mm
\end{table}

\subsection{Evaluation of clustering results}
\label{subsec:Evaluation of clustering results}

\subsubsection{Evaluation metric}
The second evaluation metric assesses clustering performance. We evaluate the results using TS2Vec (which showed the best embedding quality) combined with DBSCAN.

As shown in Table~\ref{tab:eval2}, the evaluation covers datasets from three MMORPG titles. The column labeled LLM indicates whether an LLM is applied, and eps shows the strategy for selecting the DBSCAN parameter. The value \( q \) refers to a quantile-based method from \cite{kim2025framework}, while other entries use fixed eps values without adaptive tuning.

We report \textit{access information homogeneity} (Acc\_info) as a primary metric \cite{kim2025framework,tao2019mvan}. In addition, we use three auxiliary indicators: \#Det, \textit{max average difference} (Max\_avg), and \textit{mean average difference} (Mean\_avg).

\textit{Access information homogeneity}, proposed in \cite{kim2025framework}, quantifies behavioral similarity between characters based on their login/access patterns. Lower values indicate stronger similarity, suggesting that the characters may be controlled by the same player. The minimum value of access information homogeneity is 1, as noted in \cite{kim2025framework}, \hs{and values closer to 1 are interpreted as better performance.}

Additionally, \#Det refers to the average number of bots detected per day. Max\_avg and Mean\_avg represent the maximum and mean pairwise differences in level-up intervals among all characters within the same cluster. Smaller values indicate that characters in the cluster exhibit more homogeneous progression behavior.

% \usepackage{multirow}
% \usepackage[normalem]{ulem}
\begin{table}[t]
\centering
\small
\begin{tabular}{ccccccc}
\hline
IP & LLM & $\varepsilon$ & \#Det. & Acc\_info. & Max\_avg. & Mean\_avg. \\ \hline
\multirow{5}{*}{G1} & \multirow{4}{*}{\ding{55}} & q=0.1 & 92.14 & \textit{2.66} & 32.21 & 12.35 \\
                    &                    & q=0.2 & 153.07 & 3.22 & 32.30 & 12.14 \\
                    &                    & 2.0   & 274.14 & 4.72 & 31.85 & 12.68 \\
                    &                    & 3.0   & 511.79 & 23.15 & 84.59 & 17.53 \\ \cline{2-7}
                    & \ding{51}               & q=0.1 & 49.71 & \textbf{1.80} & 25.31 & 12.55 \\ \hline
\multirow{5}{*}{G2} & \multirow{4}{*}{\ding{55}} & q=0.1 & 71.71 & \textit{2.24} & 52.35 & 22.55 \\
                    &                    & q=0.2 & 121.07 & 2.54 & 58.14 & 23.74 \\
                    &                    & 2.0   & 261.29 & 3.24 & 74.07 & 29.02 \\
                    &                    & 3.0   & 379.64 & 10.96 & 113.66 & 28.73 \\ \cline{2-7}
                    & \ding{51}               & q=0.1 & 47.07 & \textbf{1.80} & 40.83 & 19.47 \\ \hline
\multirow{5}{*}{G3} & \multirow{4}{*}{\ding{55}} & q=0.1 & 129.86 & \textit{1.81} & 43.25 & 8.60 \\
                    &                    & q=0.2 & 219.07 & 1.93 & 49.19 & 13.11 \\
                    &                    & 2.0   & 615.50 & 2.25 & 24.03 & 5.65 \\
                    &                    & 3.0   & 663.79 & 4.33 & 95.00 & 18.39 \\ \cline{2-7}
                    & \ding{51}               & q=0.1 & 84.07 & \textbf{1.41} & 13.69 & 5.32 \\ \hline
\end{tabular}
\caption{Setting $q = 0.1$ generally yields better performance. LLM refinement also significantly reduced access information homogeneity. All results are reported as daily averages.}
\label{tab:eval2}
\vskip -10mm
\end{table}

\subsubsection{Clustering results}

The results indicate that setting \( q = 0.1 \) generally leads to lower acc\_info scores across game datasets, indicating lower-risk clustering. For G3, while \( \varepsilon = 2.0 \) yielded slightly better Max\_avg and Mean\_avg scores, \( q = 0.1 \) achieved a lower acc\_info. In DBSCAN, \( \varepsilon \) controls clustering granularity: a larger value applies a looser criterion, while a smaller value results in stricter identification of suspicious accounts. As this study prioritizes minimizing false positives, we recommend \( q = 0.1 \).

While excluding such clusters is important for building a low-risk sanction list, tuning \( \varepsilon \) alone is insufficient. Thus, we introduce a double-checking mechanism using GPT-4o, with results discussed in later experiments.

\subsection{LLM-based verification}

\subsubsection{Evaluation metric}

This section evaluates the proposed method using the same metric as in Section~\ref{subsec:Evaluation of clustering results}. This metric assesses performance improvement after LLM-based refinement.

\subsubsection{Effectiveness of LLM-based refinement}

The effectiveness of LLM-based refinement is demonstrated in the rows where the LLM column is marked as used (\ding{51}) in Table 2. Experimental results show that the LLM effectively filtered out normal users from the sanction candidates in an appropriate direction, leading to a meaningful reduction in the access information homogeneity score across all three games.

% To evaluate this, we first construct ten variants of the level-up interval sequence for each character by applying different levels of perturbation. These perturbed sequences are defined as follows:

% \begin{equation}
% T^{(p)}_\text{pert} = T^{(p)} + \mathcal{N}_{lv}
% \end{equation}
% where \( \mathcal{N}_{lv} \sim \text{Uniform}(-3 \cdot lv,\; 3 \cdot lv) \) with probability \( \frac{lv}{10} \), 
% and \( lv \in \{1, 2, \dots, 10\} \).

% In addition to the formulation in Equation~(1), we further apply two operations—random deletion and index-wise swapping—to construct the perturbation term $N_{lv}$ in a more realistic manner. Given an original sequence $T^{(p)} = [t_1, t_2, \ldots, t_n]$, the perturbed sequence $\text{pert}^{(p)}_{lv}$ is defined as:
% \[
% \text{pert}^{(p)}_{lv} = \text{Swap}\left( T^{(p)}_{\text{pert},\text{del}}, \{(a_k, b_k)\}_{k=1}^{lv} \right)
% \]
% where:
% \begin{itemize}
%   \item $T^{(p)}_{\text{pert}, \text{del}} = [t_i \in T^{(p)}_{\text{pert}} \mid u_i > p_{\text{del}},\ u_i \sim \text{Uniform}(0,1)]$ \\
%   represents the result of \textit{random deletion}, with deletion probability $p_{\text{del}} = 0.05 \times lv$ for each element.
  
%   \item $\text{Swap}(\cdot)$ denotes the \textit{index-wise swapping} operation, which randomly selects $lv$ disjoint index pairs $(a_k, b_k)$ and swaps their corresponding elements.
% \end{itemize}

% This perturbation process introduces increasing levels of randomness as $lv$ increases, allowing us to evaluate the robustness and discriminative power of the representation model under structured noise.

\section{Conclusion}

This paper presents an unsupervised framework for detecting auto-leveling bots in MMORPGs using time-series representation learning and LLM-based verification. By clustering characters with similar level-up behaviors and refining the results with LLM, our method reduces labeling costs while achieving high accuracy. Experiments on real game data demonstrate its practical value for game security.

%ACKNOWLEDGMENTS are optional
% \section{Acknowledgments}
% \begin{acks}
% Proofreading support by ChatGPT is acknowledged.
% \end{acks}

\section{GenAI Usage Disclosure}
GenAI was used for proofreading only. No content was generated.
\bibliographystyle{ACM-Reference-Format}
\bibliography{samplebase}

\end{document}